# A multimodal method based on cross-attention and convolution for postoperative infection diagnosis

Xianjie Liu
College of Computer Science
Sichuan University
Chengdu, Sichuan, China
liuxianjie@stu.scu.edu.cn

Hongwei Shi*
College of Computer Science
Sichuan University
Chengdu, Sichuan, China
*Corresponding author: shihw001@126.com

*Abstract*—*Postoperative infection diagnosis is a common and serious complication that generally poses a high diagnostic challenge. This study focuses on PJI, a type of postoperative infection. X-ray examination is an imaging examination for suspected PJI patients that can evaluate joint prostheses and adjacent tissues, and detect the cause of pain. Laboratory examination data has high sensitivity and specificity and has significant potential in PJI diagnosis. In this study, we proposed a self-supervised masked autoencoder pre-training strategy and a multimodal fusion diagnostic network MED-NVC, which effectively implements the interaction between two modal features through the feature fusion network of CrossAttention. We tested our proposed method on our collected PJI dataset and evaluated its performance and feasibility through comparison and ablation experiments. The results showed that our method achieved an ACC of 94.71% and an AUC of 98.22%, which is better than the latest method and also reduces the number of parameters. Our proposed method has the potential to provide clinicians with a powerful tool for enhancing accuracy and efficiency.*

## I. INTRODUCTION

Postoperative infection refers to an infection caused by microorganisms such as bacteria or fungi after surgery. Postoperative infection brings unnecessary pain to patients and wastes a lot of medical resources [1]. Therefore, finding an effective method to diagnose postoperative infection is very important.

Multimodal technology has been widely applied in various fields. Its results have demonstrated the power of multimodal technology, which brings different advantages through the fusion of information from multiple modalities. In the medical field, this technology can also provide better diagnostic results through the interaction of information between different modalities (such as CT, MRI, physiological data, etc.) [2].

However, there are two problems in current research: firstly, the limited availability of publicly available medical multimodal data leads to poor training results of various supervised learning methods. Secondly, previous medical multimodal learning research has mainly focused on the fusion of different modalities of images or the fusion of images and natural language, with less research on the fusion of images and laboratory examination data, resulting in previous methods failing to effectively utilize laboratory examination data.

In this paper, to address these two problems, we propose a simple and effective modality fusion method for image and laboratory examination data based on self-supervised masked autoencoder (MAE), called MED-NVC (Numerical and Visual Fusion of PIJ). Our training consists of two stages: the encoder pre-trainning stage trains the visual encoder self-supervisedly by masking and restoring images, and the full data learning stage trains the entire network using laboratory examination data and images.

We collected relevant data at West China Hospital of Sichuan University, including laboratory examination data (serum C-reactive protein (CRP), erythrocyte sedimentation rate (ESR), gender, age, hypertension, diabetes, rheumatism, arthritis, anemia, osteoporosis, cerebral infarction, hypoalbuminemia, hypothyroidism, liver disease, etc.) and X-ray image data. Laboratory examination data have high sensitivity and specificity, and have significant potential in the diagnosis of postoperative infections. X-ray examination is an imaging examination for suspected postoperative infection patients, which can evaluate joint prostheses and adjacent tissues and detect the cause of pain. In early infections, X-rays show non-specific soft tissue swelling around the prosthetic joint.

## II. BACKGROUND AND RELATED WORK

### A. MAE

Masked Autoencoders Are Scalable Vision Learners (MAE) [3] is an autoencoder model that learns the hidden features of data by encoding it into a low-dimensional vector and then decoding it back to the original data. Unlike traditional autoencoders, MAE masks part of the input data and only learns the features of the unmasked data, thus improving learning efficiency and generalization ability. MAE is widely used in tasks such as image classification, object detection, and image generation, and has achieved good results in practice.

### B. ConvNeXt-V2

ConvNeXt-V2 [4] is a novel convolutional neural network model that combines self-supervised learning and new architectural improvements to improve the efficiency of image recognition. ConvNeXt-V2 uses a fully convolutional masked autoencoder (FCMAE) and global response normalization (GRN) to enhance feature competition between channels.

*C. Cross-Attention*

The Multi-Modality Cross Attention Network for Image and Sentence Matching (MMCA) [5] proposed a multi-modality cross-attention mechanism to facilitate multi-modal matching between images and text. However, it only focused on multi-modal matching between natural language and images, and did not investigate digital data such as laboratory examination data. Therefore, we propose a new model structure approach (called MED-NVC-Net) based on MAE, which maps laboratory examination data to the same dimension as patch-based image data and then uses cross-attention to fuse the two data modalities. Finally, we use the CrossEntropy loss function for classification to better guide the model in distinguishing medical cases.

## III. METHOD

The MAE-NVC network architecture we propose utilizes self-supervised Masked Autoencoder (MAE) and a designed multimodal information interaction method. The entire network structure is shown in Figure 1. The network is divided into two-stage training. In the encoder pre-training stage, we train the image encoder through self-supervised learning by randomly masking 60% of the images and feeding the remaining images into the image encoder. Then, the encoder's features are used to reconstruct the images for self-supervised training. In the full data learning stage, we input complete images and laboratory examination data. The images are first processed by the image encoder's first two stages and then inputted into the one-way CrossAttention along with up-sampled laboratory examination data. Finally, a set of multimodal fused features is obtained, which is used for the final classification.

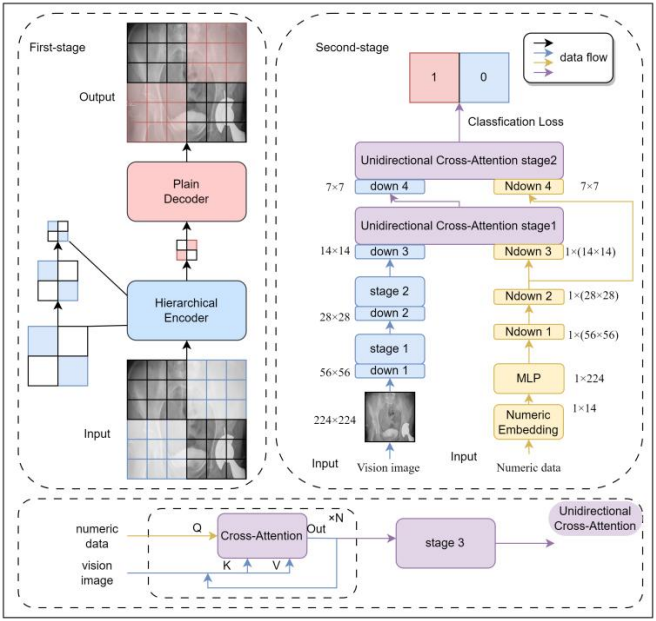

Figure 1. The architecture of the MAE-NVC network. The training is conducted in two stages. In the encoder pre-training

Sichuan Provincial student Innovation and Entrepreneurship Funding:C2023125222

stage, the image encoder is self-supervised trained by masking. In the full data learning stage, supervised training is performed.

*A. Encoder pre-training*

In the initial pre-training stage of the image encoder, the model aims to capture the internal information of the images themselves. When designing the Encoder, since MAE is implemented based on the Transformer model, visible image blocks can be the only input to the model. However, it is difficult to preserve the 2D image structure in ConvNets. A simple way is to directly mask the image [6,7], but this reduces learning efficiency and leads to inconsistency between training and testing because there are no mask labels during testing. This becomes a serious problem when the masking rate is high.

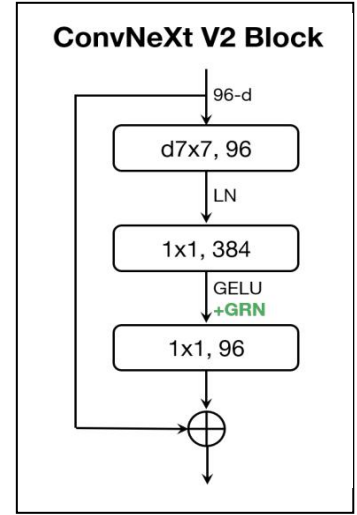

Figure 2. A ConvNeXt-V2 block is used in this paper, which incorporates a GRN layer [11] after the dimension-expansion MLP layer and drops LayerScale. It is employed in both the image encoder and decoder.

We refer to the idea of ConvNeXt-V2: the masked image can be represented as a 2D sparse pixel array [8,9,10]. Naturally, we use sparse convolution layers, which allow the model to operate only on visible images. Moreover, note that sparse convolution layers can be easily adjusted back to standard convolution in the full data learning stage. Finally, we use ConvNeXt-v2 block as Figure 2. in our encoder.

Following the MAE idea and considering consistency, we do not use the Transformer but design a simple and lightweight ConvNeXt-V2 block as the Decoder, forming an asymmetric Encoder-Decoder structure. Similar to ConvNeXt-V2, we set the dimension of the Decoder to 512.

We designed a reconstruction target: calculating the mean squared error (MSE) between the reconstructed image and the target image. Similar to MAE, the target is the block-normalized image of the original input, and the loss is only applied to the masked patches.

After the entire network is built, we use parts of the images in the dataset to pre-train the image Encoder. In the encoder pre-training stage, the model is trained on a single NVIDIA RTX 4090 GPU with a batch size of 24 for a total of 800 epochs. The Lion optimizer with default parameters is used.

The learning rate is warmed up to 5e-5 in the first 10 epochs, then decays according to the cosine schedule to 5e-7 [12]. Images are adjusted to a resolution of 224×224, with random horizontal flipping and rotation applied for augmentation. To speed up the training process, the NVIDIA AMP strategy is used.

## B. Full data learning

In the Full data learning stage, the model needs to undergo supervised training with labeled data. At this stage, the model's input becomes two modalities: images and laboratory examination data. Images do not need to be masked at this stage.

First, the images are processed by the image Encoder pretrained in the first phase to extract features. In order to perform CrossAttention between the laboratory examination data and the image data, we adjust the dimensionality of the laboratory examination data to match that of the image data. To enable interaction between the image features and the other modality, we design a modality fusion network. In this network, images are used as Q, and the laboratory examination data are used as K and V for the following calculation:

$$\text{Cross\_Attention}(Q^V, K^N, V^N) = \text{softmax}(\frac{QK^T}{\sqrt{d_k}})V \quad (1)$$

In Unidirectional Cross-Attention, the resulting calculation will be input into the next block as Q or the output of the entire stage. In other words, in the Unidirectional Cross-Attention stage, only the image data has been changed, while the laboratory examination data remains unchanged. After performing cross-attention, the resulting calculation will be sent to the image encoder pre-trained in the encoder pre-training stage to extract features. Finally, we use Cross-Entropy Loss to calculate the loss between the classification result and the target. The full data learning stage model was trained using a single NVIDIA RTX 4090 GPU with a batch size of 24 for 160 epochs. We used the Lion optimizer with default parameters. The learning rate was warmed up to 5e-6 for the first 10 epochs and then decayed according to a cosine annealing schedule to 5e-8. The images were adjusted to a resolution of 224×224, and data augmentation techniques such as random horizontal flipping and rotation were applied. The size of the laboratory examination data vector was set to 1×14. During training, to enable the model to identify a single modality separately, we randomly replaced the numerical values of a single modality vector with -10 with a probability of 0.2. To speed up the training process, we used the NVIDIA AMP strategy.

## C. Training Loss

The loss function for the encoder pre-training stage uses MSELoss:

$$L = \text{MSELoss}(f_{out}, y) \quad (2)$$

In this equation, y represents the chunk-normalized image of the original input, and the loss function used is MSELoss.

The loss function for the full data learning stage uses CrossEntropyLoss:

$$L = \text{CrossEntropyLoss}(f_{out}, \text{lable}) \quad (3)$$

In this equation, lable represents the chunk-normalized image of the original input, and the loss function used is CrossEntropyLoss.

## D. Evaluation Metrics

Accuracy (ACC) is a widely used performance metric that measures the proportion of correct predictions relative to the total number of predictions made by a model. Mathematically, it can be represented as:

$$ACC = \frac{TP+TN}{TP+TN+FP+FN} \quad (4)$$

Where TP is the number of true positive cases, TN is the number of true negative cases, FP is the number of false positive cases, and FN is the number of false negative cases.

The area under the curve (AUC) is a performance metric used to evaluate the trade-off between the true positive rate (recall or sensitivity) and the false positive rate (1-specificity) at different threshold settings.

## IV. EXPERIMENTS

### A. Dataset

In this article, we collected relevant data from West China Hospital of Sichuan University, including 221 positive patients and 268 negative patients. The dataset includes 1,170 images and 489 laboratory examination data. Each image was cropped to 224×224, and each laboratory examination data has a length of 14. An example of a data shown as Table 1. and Figure 3.

TABLE I.     AN EXAMPLE OF A DATA SET IS SHOWN BELOW (DATA HAS BEEN ANONYMIZED TO PROTECT PATIENT PRIVACYS).

| TYPE | numerical value |
|---|---|
| CRP | 1.24 |
| ESR | 8 |
| Hip or knee | hip |
| Position | 1 |
| Sex | 0 |
| Age | 62 |
| Hypertension | 0 |
| Diabetes | 0 |
| Rheumatoid Arthritis | 0 |
| Anemia | 0 |
| Osteoporosis | 0 |
| Cerebral infarction | 0 |
| Hypoalbuminemia | 0 |
| Hypothyroidism | 0 |
| Liver Disease | 0 |

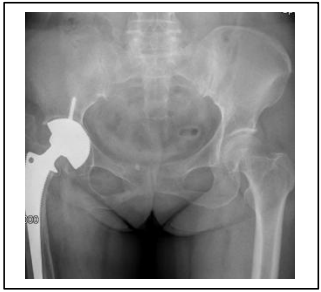

*Figure 3. A vision example of a data.*

TABLE II. Performance comparison under the dataset configuration.

| Model | Eval size | Multimodal Fusion | Parameters | AUC | ACC |
|---|---|---|---|---|---|
| MLP | 224² | Only-numric | 0.016M | 92.12% | 84.96% |
| ConvNeXt-V2-B | 224² | Only-vision | 83.63M | 85.93% | 82.52% |
| ConvNeXt-V2-B | 224² | MLP | 83.87M | 94.95% | 91.46% |
| Swin-B | 224² | MLP | 82.73M | 96.16% | 89.02% |
| MAE-NVC-without pretrain | 224² | Unidirectional Cross-Attention | 77.26M | 92.65% | 88.21% |
| MAE-NVC -no random throw | 224² | Unidirectional Cross-Attention | 77.26M | 97.96% | 93.08% |
| **MAE-NVC** | **224²** | **Unidirectional Cross-Attention** | **77.26M** | **98.22%** | **94.71%** |

### B. Experiment on Dataset

To evaluate the performance of our model, we conducted experiments on the PJI dataset using various multimodal methods. We compare our results to the state-of-the-art architecture designs, including ConvNeXt-V2-Base, SwinTransformer-Base [13].

In the visual modality network, laboratory examination data passed through a three-layer multilayer perceptron (MLP) with GELU [14] activation function processing, while images were processed through the entire network, accompanied by Maxpool operations in the batch dimension. The fused image and text features were directly concatenated for classification prediction.ConvNeXt-V2-B and Swin-B both used pretrain and random throw. Under the basic setup, the MAE-NVC model outperformed the state-of-the-art models significantly as Table 2. The MAE-NVC achieved a Top-1 accuracy of **94.71%** and an AUC of **98.22%**. Outperforming the strongest model by **3.25%** Top-1 accuracy and **2.06%** AUC, while using fewer parameters.

In conclusion, our proposed method demonstrates significant advantages by utilizing self-supervised masked autoencoder pre-training and adopting Unidirectional Cross-Attention for modality fusion.

## V. Conclusions

This article introduces an innovative solution for postoperative infection detection, which improves diagnostic accuracy by using X-Ray images and laboratory examination data. The self-supervised masked autoencoder pre-training effectively mines the features of the images. In addition, the Unidirectional Cross-Attention Block is used to achieve deep fusion of multiple modalities. This method is validated using a self-collected dataset of postoperative infections and compared with various methods under the same experimental configuration. The results show that the ACC of this method reaches **94.71%** and the AUC reaches **98.22%**. In the future, our goal is to combine this method with the capabilities of large-scale language models to achieve automatic diagnosis and diagnostic text writing for multiple comorbidities.

## References

[1] O'Brien W J, Gupta K, Itani K M F. Association of postoperative infection with risk of long-term infection and mortality[J]. JAMA surgery, 2020, 155(1): 61-68.

[2] Kumar A, Kim J, Cai W, et al. Content-based medical image retrieval: a survey of applications to multidimensional and multimodality data[J]. Journal of digital imaging, 2013, 26: 1025-1039.

[3] Kaiming He, Xinlei Chen, Saining Xie, Yanghao Li, Piotr Dollár, Ross Girshick. Masked Autoencoders Are Scalable Vision Learners. arXiv:2111.06377, 2021.

[4] Sanghyun Woo, Shoubhik Debnath, Ronghang Hu, Xinlei Chen, Zhuang Liu, In So Kweon, Saining Xie. ConvNeXt V2: Co-designing and Scaling ConvNets with Masked Autoencoders. arXiv:2301.00808, 2023.

[5] X. Wei, T. Zhang, Y. Li, Y. Zhang and F. Wu, "Multi-Modality Cross Attention Network for Image and Sentence Matching," 2020 IEEE/CVF Conference on Computer Vision and Pattern Recognition (CVPR), Seattle, WA, USA, 2020, pp. 10938-10947, doi: 10.1109/CVPR42600.2020.01095.

[6] Hangbo Bao, Li Dong, and Furu Wei. BEiT: BERT pretraining of image transformers. In ICLR, 2022.

[7] Zhenda Xie, Zheng Zhang, Yue Cao, Yutong Lin, Jianmin Bao, Zhuliang Yao, Qi Dai, and Han Hu. Simmim: A simple framework for masked image modeling. In CVPR, 2022.

[8] Christopher Choy, JunYoung Gwak, and Silvio Savarese. 4d spatio-temporal convnets: Minkowski convolutional neural networks. In CVPR, 2019.

[9] Benjamin Graham, Martin Engelcke, and Laurens Van Der Maaten. 3d semantic segmentation with submanifold sparse convolutional networks. In CVPR, 2018.

[10] Benjamin Graham and Laurens van der Maaten. Submanifold sparse convolutional networks. arXiv preprint arXiv:1706.01307, 2017.

[11] Hugo Touvron, Matthieu Cord, Alexandre Sablayrolles, Gabriel Synnaeve, and Herve Jégou. Going deeper with image transformers. In ICCV, 2021

[12] Loshchilov, I.; Hutter, F. Fixing weight decay regularization in adam 2017.

[13] Ze Liu, Yutong Lin, Yue Cao, Han Hu, Yixuan Wei, Zheng Zhang, Stephen Lin, Baining Guo. Swin Transformer: Hierarchical Vision Transformer using Shifted Windows. arXiv:2103.14030, 2021.

[14] Gardner, M.W.; Dorling, S. Artificial neural networks (the multilayer perceptron)—a review of applications in the atmospheric sciences. Atmospheric environment 1998, 32, 2627–2636.